\documentclass[sigconf,review]{acmart}

\usepackage{booktabs}
\usepackage{subcaption}
\usepackage{graphicx}
\usepackage{tikz}
\usetikzlibrary{arrows.meta,positioning,fit,backgrounds}
\usepackage{pgfplots}
\pgfplotsset{compat=1.18}
\usepackage{xcolor}
\usepackage{CJKutf8}

\newcommand{\Clock}{\textsf{Clock}}

\setcopyright{none}
\acmDOI{}
\acmISBN{}
\renewcommand\footnotetextcopyrightpermission[1]{}
\begin{document}

\title[Can Agents Keep Up with Time? An Empirical Analysis of Temporal Adaptation After a Conversational Hiatus]{Can Agents Keep Up with Time?\texorpdfstring{\\}{ }An Empirical Analysis of Temporal Adaptation After a Conversational Hiatus}

\author{Hongguan Su}
\authornote{These authors contributed equally to this work.}
\affiliation{%
  \institution{People's Public Security University of China}
  \city{Beijing}
  \country{China}
}

\author{Ming Wang}
\authornotemark[1]
\affiliation{%
  \institution{Northeastern University}
  \city{Shenyang}
  \country{China}
}

\author{Shun Zhang}
\affiliation{%
  \institution{People's Public Security University of China}
  \city{Beijing}
  \country{China}
}

\author{Yifan Wang}
\affiliation{%
  \institution{People's Public Security University of China}
  \city{Beijing}
  \country{China}
}

\author{Rong Yin}
\affiliation{%
  \institution{Beihang University}
  \city{Beijing}
  \country{China}
}

\author{Zhiming Ma}
\affiliation{%
  \institution{JD.com}
  \city{Beijing}
  \country{China}
}

\author{Mengzheng Yang}
\affiliation{%
  \institution{Henan University}
  \city{Kaifeng}
  \country{China}
}

\author{Peng Chen}
\affiliation{%
  \institution{People's Public Security University of China}
  \city{Beijing}
  \country{China}
}

\begin{abstract}
Long-horizon agents increasingly serve as persistent personal assistants across modern web platforms, managing schedules, applications, and multi-step plans over weeks or months. Users often return to an ongoing conversation after a long conversational hiatus, while earlier bookings and preparation notes stay on record. During this hiatus, scheduled dates may pass even though the retained records remain unchanged and event outcomes go unrecorded. Existing evaluations of long-horizon agents focus on fact retention and explicit state updates. Whether agents perceive the passage of time and adapt their advice accordingly therefore remains unknown. We frame this ability as temporal state perception and use, which means perceiving the current time and applying it when interpreting dated records. To study it, we design session-rekindling tasks that keep the earlier conversation fixed and vary only the follow-up date and the way temporal information reaches the agent. Experiments across five model routes and five temporal-information conditions show that agents can obtain the current time but often fail to use it. They reliably query the clock and even note in their reasoning that the scheduled date has passed, yet many still tell users to prepare for the expired event. Supplying explicit temporal context reduces this error, and providing the computed date relation removes it entirely. Keeping up with time therefore depends on turning perceived time into updated plans. Based on these findings, we design a session-rekindling benchmark that provides temporal-awareness guidance for designing long-horizon agents. The code can be found at \url{https://github.com/Adoreu2/session-rekindling-benchmark}.
\end{abstract}



\maketitle
\pagestyle{plain}
\let\balance\relax

\section{Introduction}

\begin{figure}[t]
\centering
\includegraphics[width=\columnwidth]{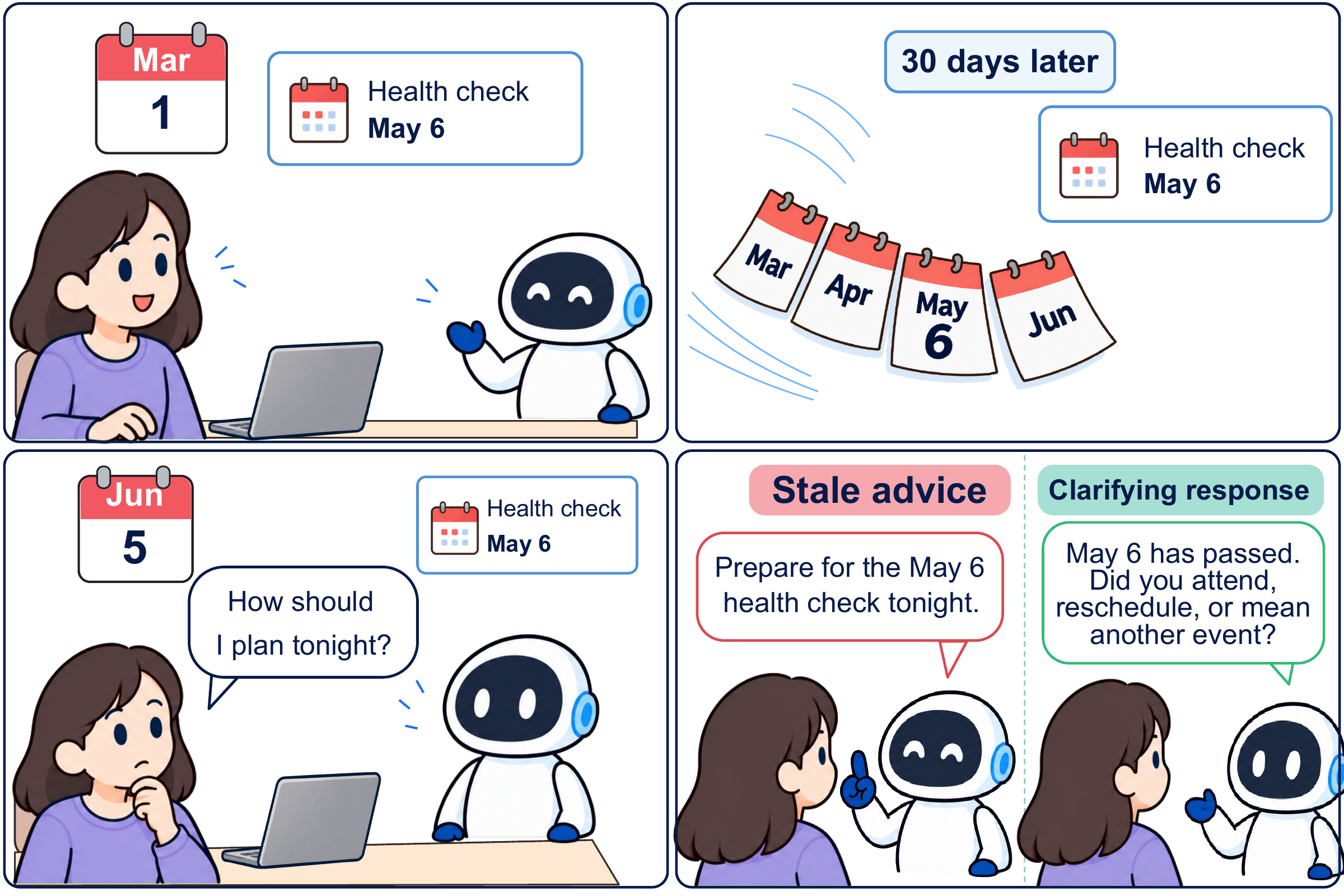}
\caption{A session-rekindling scenario where the scheduled May~6 health check has passed by the June~5 follow-up, contrasting stale preparation advice with an appropriate clarifying response.}
\Description{A timeline figure: an earlier session books a May~6 health check, the user returns on June~5 asking how to plan the evening, and two contrasting responses appear side by side---stale preparation advice and a clarifying question about whether the appointment took place.}
\label{fig:timeline}
\end{figure}

Consider Mary who books a medical check-up for May~6 through a web-based conversational assistant, as illustrated in Figure~\ref{fig:timeline}. Notes from the initial interaction record the appointment time alongside instructions to fast beforehand, bring confirmation documents, and rest early. Weeks later, on June~5, the user returns to the same conversation thread and asks how to plan for the evening. On May~5, one day before the scheduled examination, the assistant should naturally provide dietary and preparation guidelines for the next morning. By June~5, the scheduled date has long passed, and the retained record leaves open whether the user attended, missed, or rescheduled the appointment. An assistant that overlooks this elapsed calendar time prepares the user for an outdated event, misinterpreting user intent and offering obsolete advice.

This failure exposes a fundamental blind spot in long-horizon agents deployed as persistent assistants across modern web services. As conversational interfaces increasingly mediate healthcare scheduling, travel planning, visa applications, and enterprise workflows, user interactions naturally span multiple weeks or months. While these systems manage tasks across extended periods, the external world continues to advance even when conversation logs and stored memory items remain static. Existing evaluations of long-horizon agents examine cross-session fact retention, temporal tool calls, and memory revision under explicit updates~\cite{maharana2024longterm,wu2025longmemeval,cheng2025temporally,chao2026stale}. How agents perceive the silent passage of calendar time and adapt their recommendations when earlier outcomes remain unresolved is still unknown. Because the stored schedule itself stays unchanged during a conversational hiatus, an agent can accurately retrieve historical notes yet answer for a moment that has already passed.

We define this target capability as \emph{temporal state perception and use}, which requires perceiving the current reference time and integrating it with outcome uncertainty when interpreting dated records. When a user returns to an ongoing conversation after a conversational hiatus, a process we term \emph{session rekindling}, the agent must determine whether stored plans remain applicable. A dated booking record remains historically accurate for retrospective queries, yet its prospective status shifts from upcoming to passed once the reference date crosses the scheduled boundary. Because the conversation log lacks an outcome record, the agent must also recognize that attendance or completion remains unknown. Three research questions organize our investigation:

\begin{itemize}
  \item \textbf{RQ1:} Can agents recognize that a task is time-sensitive?
  \item \textbf{RQ2:} Do agents update the task state as time passes?
  \item \textbf{RQ3:} Is the update correct, and does the advice fit the situation?
\end{itemize}

\begin{figure*}[t]
\centering
\includegraphics[width=\textwidth]{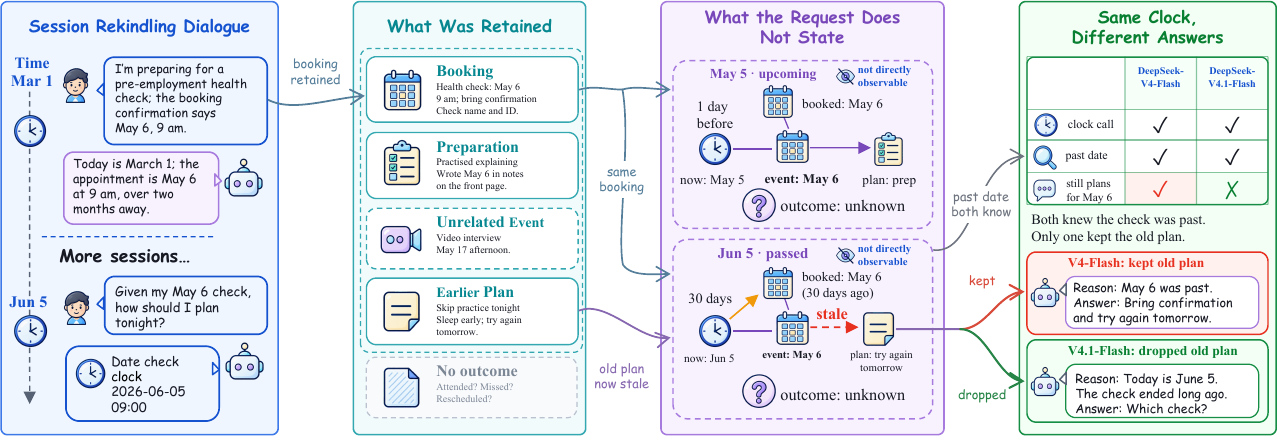}
\caption{Structure of a session-rekindling episode re-engaged on June~5, comparing retained records, unobserved event outcomes, and divergent responses from two model routes under identical clock readings.}
\Description{A four-panel episode figure: the earlier dialogue and the clock call, the retained materials with no outcome record, the upcoming-versus-passed states the request does not state, and two runs under the same clock result---DeepSeek-V4-Flash keeps the stale preparation plan whereas DeepSeek-V4.1-Flash drops it and asks which event is meant.}
\label{fig:comic}
\end{figure*}

To answer these questions, we design \emph{session-rekindling tasks}, which are controlled service episodes where a user resumes an ongoing conversation after a conversational hiatus (Figure~\ref{fig:comic}). Across each episode, the historical dialogue prefix, preparation notes, and user plans remain fixed while we advance a virtual clock tool named \Clock{} across pre-date and post-date offsets. Keeping the retained context and follow-up request identical across time points ensures that any required shift in advice stems purely from elapsed time. We further vary how temporal evidence reaches the agent across five experimental conditions ranging from omitting the date to supplying explicit date relations. Using this controlled harness, we collect 940 responses across five model routes and apply rule matchers alongside a condition-blind dual-judge protocol. This evaluation pipeline separately measures active clock invocation, past-date recognition in exposed reasoning, schedule framing in the final response, and overall advice acceptability.

Our experiments across the five model routes reveal a sharp dissociation between acquiring the current time and applying it to update final advice. On RQ1, agents recognize temporal dependencies when prompts omit the date and reliably invoke the \Clock{} tool to fetch the current time. On RQ2, acquiring the timestamp rarely guarantees updating the plan, as post-date responses frequently continue to frame elapsed schedules as upcoming. Exposed reasoning traces often state explicitly that the scheduled date has passed while the final answer still instructs the user to prepare for the expired appointment. Supplying explicit date relations eliminates this upcoming framing completely, pinpointing temporal state deduction as the primary bottleneck. On RQ3, responses that retain upcoming framing fail the advice rubric entirely, whereas dropping upcoming framing often triggers unwarranted completion assumptions. Building on these findings, we consolidate our task episodes and scoring protocol into a diagnostic session-rekindling benchmark and make three primary contributions:

\begin{itemize}
  \item We formalize temporal state perception and use for long-horizon agents, establishing an analytical framework that separates calendar progression from evidence about event outcomes during session rekindling.
  \item We conduct controlled experiments across five model routes and five temporal conditions, demonstrating that active clock retrieval and past-date reasoning frequently dissociate from final response framing.
  \item We construct and release a diagnostic session-rekindling benchmark paired with a validated dual-judge evaluation protocol to guide the design of temporally aware long-horizon agents.
\end{itemize}

\section{Related Work}

\subsection{Evaluating Web-facing and long-horizon agents}

Benchmarks for web-facing and conversational agents evaluate diverse interaction capabilities across static snapshots and live execution environments. Mind2Web uses recorded webpage states and annotated action sequences to evaluate element selection and operation prediction at each step~\cite{deng2023mind2web}. WebLINX studies multi-turn conversational browser navigation where user instructions evolve across turns, comparing predicted agent actions against human demonstrations~\cite{lu2024weblinx}. In interactive execution environments, WebArena evaluates functional task completion through programmatic state checks~\cite{zhou2024webarena}, while WorkArena measures enterprise workflow execution across knowledge-work interfaces~\cite{drouin2024workarena}. TheAgentCompany extends interactive evaluation to consequential workplace tasks using milestone checkpoints~\cite{xu2025agentcompany}, and BrowserGym unifies web-agent evaluation interfaces across benchmarks~\cite{lesellier2025browsergym}. AssistantBench focuses on time-consuming web tasks with verifiable answers while filtering out queries whose ground-truth answers shift rapidly over time~\cite{yoran2024assistantbench}. Alongside navigation suites, long-term conversational memory benchmarks test whether assistants accurately retrieve facts across extended multi-session histories~\cite{maharana2024longterm,wu2025longmemeval}. Our study complements both lines of research by holding retrieved records constant and testing whether the agent adapts its advice when the reference date advances past a scheduled event.

\subsection{Temporal adaptation and expiring state in multi-session interactions}

Recent benchmarks have begun to examine how language agents handle timestamps, prospective intentions, and state transitions across multi-session interactions. TIMEBench evaluates temporal-adaptivity scenarios where deadlines expire and assumptions grow stale over a conversational hiatus, crediting awareness in either reasoning traces or final replies~\cite{das2026time}. TicToc holds prior dialogue timestamps fixed while varying the hiatus duration before a follow-up query, scoring tool invocation decisions against human temporal preferences~\cite{cheng2025temporally}. In parallel, PM-Bench and TriggerBench evaluate prospective memory by testing whether agents trigger time-based intentions when a monitored clock enters a valid window~\cite{liu2026pmbench,zhang2026trigger}. ContractBench audits observation contracts including temporal expiry in execution traces using deterministic validators~\cite{wang2026contract}. Meanwhile, STALE measures memory reuse and planning adaptation after new conversational evidence invalidates earlier records~\cite{chao2026stale}. StateMemBench tracks state transitions and expiration across evolving logs~\cite{fan2026track}, and StateAuditor examines implicit stale dependencies following explicit memory updates~\cite{sun2026repair}. Whereas those settings either score tool decisions or update memory through new textual observations, our framework isolates response-level adaptation to silent calendar progression under open outcome uncertainty.

\subsection{Exposed reasoning and cognitive perspectives on temporal adaptation}

Prior studies on temporal reasoning and chain-of-thought faithfulness provide important context for analyzing how models process dated evidence. TimeAware benchmarks dated fact recall across shifting reference periods, requiring models to compare recorded dates against query timestamps~\cite{herel2025timeaware}. In parallel, research on reasoning faithfulness examines whether intermediate chain-of-thought steps genuinely govern final model predictions under targeted interventions~\cite{lanham2023faithfulness}. Our evaluation bridges these perspectives at the response level by separately scoring past-date recognition in exposed reasoning and schedule framing in the final advice. This separation parallels findings from human prospective memory research on intention deactivation~\cite{moschl2020aftereffects}. In cognitive psychology, individuals who fail to deactivate a completed or expired intention commit commission errors by repeating obsolete actions after an intention has expired. Furthermore, natural language comprehension requires shifting the deictic center of relative temporal markers such as \emph{tonight} and \emph{tomorrow} as reference time advances. Session rekindling combines both cognitive demands because the passage of a scheduled date invalidates preparation plans while leaving the actual event outcome unknown.

\section{Problem Formulation}
\label{sec:formulation}

\subsection{Agent inputs and the required state}

A session-rekindling task is an interaction episode in which a user returns to an ongoing conversation after a conversational hiatus to seek follow-up advice on a scheduled goal. During this return, termed session rekindling, the agent relies on retained session records that have remained static since the earlier exchange. In each task instance, the agent receives a retained record $m$ specifying an event scheduled for date $d(m)$ and a follow-up question $q$. The agent also receives access to the virtual clock tool \Clock{} and outcome evidence $E$. Throughout our controlled setting, the evidence set $E$ contains only the initial schedule, leaving event completion, cancellation, or rescheduling open. Let $\kappa$ denote the environmental current time supplied by the harness and let $r$ denote an explicit reference date specified by the user. The effective reference time is $t^{*}=r$ when the query requests advice as of a named date and $t^{*}=\kappa$ in standard interactions.

Although the booking record $m$ remains historically accurate at every point in time, its appropriate application depends on the interplay between the query $q$ and the reference time $t^{*}$. A retrospective question asking when the appointment was originally booked can always be answered directly from $d(m)$. When the follow-up query $q$ instead solicits planning advice for the current evening, the relevance of the stored preparation checklist depends on whether $d(m)$ still lies ahead. The \emph{required state} $\sigma_{\mathrm{req}}(m,q,t^{*},E)$ represents the temporal interpretation of the retained schedule that the response should adopt. Here $q$ solicits planning advice while $E$ leaves the outcome open, so the required state depends directly on the relation between $t^{*}$ and $d(m)$. For any evaluation date $t^{*}\neq d(m)$, the required state is defined as follows:
\begin{equation}
\sigma_{\mathrm{req}}(t^{*},m)=
\begin{cases}
\text{upcoming}, & t^{*}<d(m),\\[2pt]
\text{passed-unresolved}, & t^{*}>d(m).
\end{cases}
\label{eq:req}
\end{equation}

A required state is \emph{time-driven} when advancing the reference time $t^{*}$ across $d(m)$ transitions $\sigma_{\mathrm{req}}$ from upcoming to passed-unresolved while $m$, $q$, and $E$ remain invariant. Before the event date ($t^{*}<d(m)$), the agent should treat the appointment as upcoming and assist with preparation. Once the event date has passed ($t^{*}>d(m)$), the agent must transition to the passed-unresolved state, deactivate obsolete preparation instructions, and account for the unrecorded outcome. This formulation generalizes across date-sensitive service domains, including medical examinations, visa interviews, qualification tests, and administrative appointments. Holding the textual inputs constant across evaluation dates isolates whether an agent updates its interpretation of a static record purely in response to elapsed time.

\subsection{From questions to observables}
\label{sec:observables}

Table~\ref{tab:rqmap} maps each research question to concrete behavioral observables and failure criteria. Formally, RQ1 asks whether the agent actively recognizes the temporal dependency and acquires the date when the follow-up request omits the current time. The indicator $F$ records whether the agent calls the \Clock{} tool during the response, and comparing $F$ across Natural and Delivered conditions characterizes selective tool acquisition. RQ2 asks whether the agent treats the event as past once the scheduled date has passed ($t^{*}>d(m)$). It also examines whether upcoming framing persists even after exposed reasoning states that the date has passed, and whether supplying explicit temporal evidence corrects this error. The primary judge label $S$ captures upcoming framing in the final response. A value of $S{=}1$ is expected pre-date ($t^{*}<d(m)$) and counts as a state-update failure post-date, while matched comparisons across conditions measure how supplied evidence alters $S$.

To probe internal state tracking under RQ2, we also examine the agent's intermediate chain-of-thought text. The rule indicator $R$ records whether exposed reasoning explicitly states that the scheduled date has passed. Even when $R{=}1$, an agent may fail to commit that deduction to its final advice. Comparing exposed reasoning against final response framing yields the post-date co-occurrence diagnostic $U$:
\begin{equation}
U=\mathbf{1}\{R{=}1 \wedge S{=}1\},\qquad t^{*}>d(m).
\label{eq:cooc}
\end{equation}

A value of $U{=}1$ identifies responses where exposed reasoning acknowledges that the event date has passed while the final answer still frames the schedule as upcoming. We complement $R$ with an independent answer rule $A$ that flags explicit past-date statements in the final response text itself. Formally, RQ3 asks whether the updated state is correct and whether the advice logically fits the passed-unresolved state implied by the elapsed date. The rubric acceptance label $P$ scores this fit, supported by two auxiliary indicators. The judge label $T$ records whether the final answer places the scheduled date in the past, and the lexical flag $K$ marks completion-associated wording in the response. Together, these observables separate tool acquisition ($F$), intermediate temporal recognition ($R$), schedule framing ($S$, $U$), and advice appropriateness ($P$, $T$, $K$).

\begin{table}[t]
  \centering\small
  \caption{Mapping from research questions to behavioral observables, expected post-date outcomes, and failure criteria.}
  \label{tab:rqmap}
  \setlength{\tabcolsep}{4pt}
  \begin{tabular}{@{}p{0.07\columnwidth}p{0.27\columnwidth}p{0.36\columnwidth}p{0.16\columnwidth}@{}}
    \toprule
    RQ & Observables & Expected behavior & Counted failure \\
    \midrule
    RQ1 & $F$ and explicit-date baseline & Clock calls track missing time cues & Rate measure \\
    RQ2 & $S$, $R$, and co-occurrence $U$ & $S{=}0$ post-date and drop under interventions & $S{=}1$ post-date \\
    RQ3 & $P$ with auxiliary $T$ and $K$ & $P{=}1$ post-date & $P{=}0$ post-date \\
    \bottomrule
  \end{tabular}
\end{table}

\section{Experimental Design}

\subsection{Episode materials and controlled time}

\begin{figure*}[t]
\centering
\includegraphics[width=0.8\textwidth]{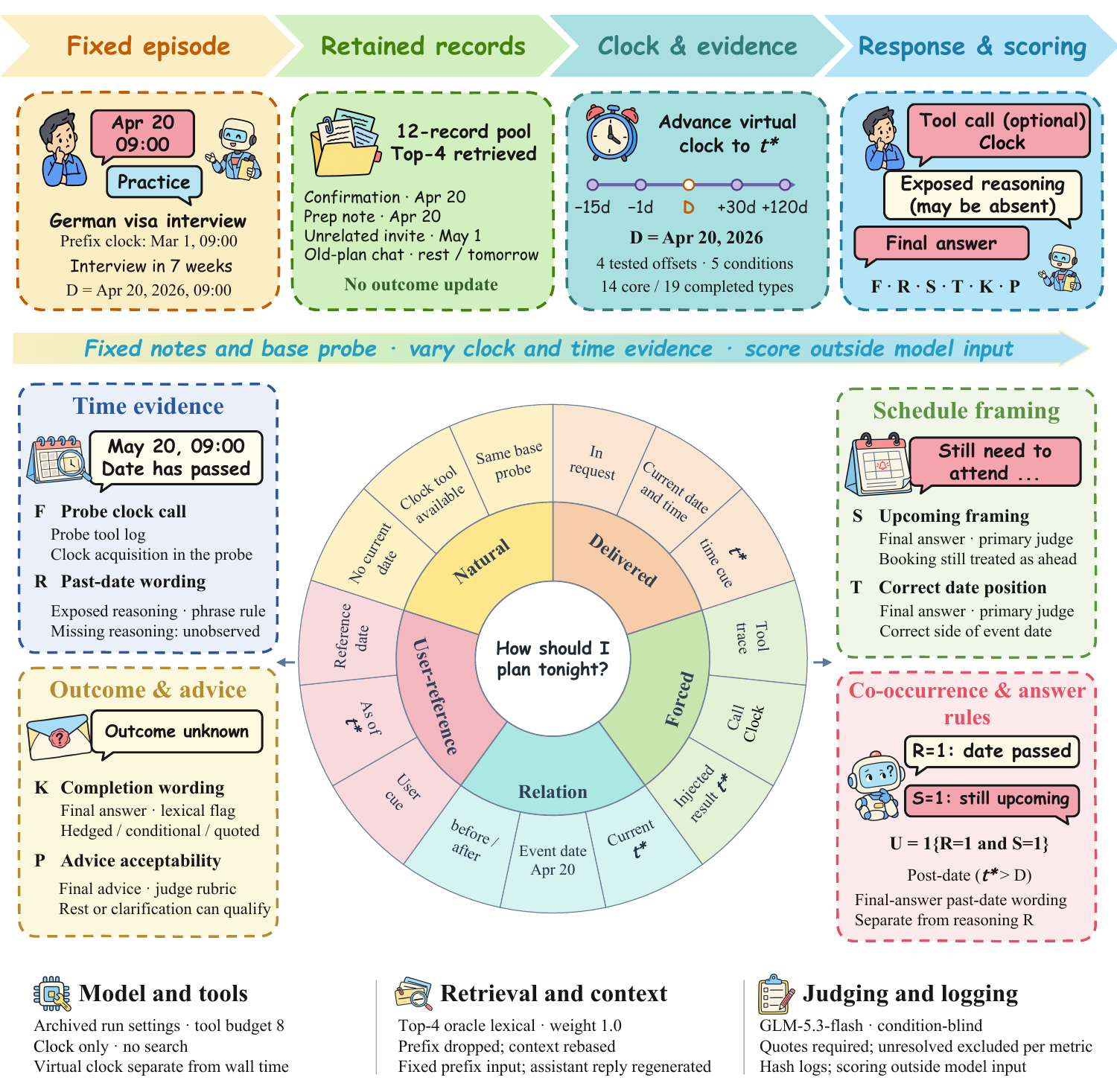}
\caption{Evaluation framework showing the fixed episode pipeline, the five temporal-information conditions around the base probe, and the response-level observables.}
\Description{A framework diagram: fixed episode materials flow through retained records and a clock probe to five temporal-information conditions, then to response scoring with the eight observables.}
\label{fig:framework}
\end{figure*}

Figure~\ref{fig:framework} illustrates our controlled evaluation framework and scoring pipeline. Each session-rekindling episode is constructed from a structured service profile specifying a scheduled event and its booking date $D$, using one profile per event type. Every episode includes the target scheduled event, preparation guidelines, and an unrelated secondary invitation on a different date to verify event grounding. The retained context also records the user's earlier plan to rest tonight and resume preparation the next day. All task materials are synthetic and written in Chinese, beginning with a standardized dialogue prefix that establishes the initial booking context on March~1, 2026. The harness pairs this fixed history with the retained notes and advances the virtual clock to four offsets ($d{-}15$, $d{-}1$, $d{+}30$, and $d{+}120$ days). Each episode retrieves the top four records from a 12-record pool via oracle lexical matching, and the agent may invoke the \Clock{} tool up to eight times per turn.

\subsection{Temporal-information conditions}

We design five conditions that systematically vary how temporal evidence reaches the agent, summarized in Table~\ref{tab:conditions}. Under the Natural condition, the follow-up prompt asks the user's open-ended planning question while omitting the current date, requiring the agent to call \Clock{} to discover the reference time. Within each route and profile, the prompt text is byte-identical across all evaluation dates, so matched comparisons isolate the behavioral impact of elapsed calendar time. Relative to Natural, the remaining four conditions progressively externalize the steps required for temporal adaptation.

The Delivered condition embeds the current timestamp directly into the user request at $d{-}1$, $d{+}30$, and $d{+}120$, testing whether prompt-level visibility reduces stale framing. The Forced condition injects a completed \Clock{} call and its returned timestamp into the trace before generation at $d{+}30$ and $d{+}120$, removing tool-call variance. The Relation condition provides a structured time-context observation stating the current date, the scheduled event date, and whether the event date has passed at $d{+}30$ and $d{+}120$. Finally, the User-reference condition instructs the agent to answer the query as of a named reference date ($d{-}15$ or $d{+}30$), with the virtual clock matching that date.

\begin{table}[t]
  \centering\small
  \caption{Summary of the five temporal-information conditions and their evaluated date offsets.}
  \label{tab:conditions}
  \setlength{\tabcolsep}{4pt}
  \begin{tabular}{@{}lll@{}}
    \toprule
    Condition & Temporal input & Evaluation dates \\
    \midrule
    Natural        & \Clock{} tool available & $d{-}1$, $d{+}30$, $d{+}120$ \\
    Delivered      & Current date in prompt  & $d{-}1$, $d{+}30$, $d{+}120$ \\
    Forced         & \Clock{} result injected & $d{+}30$, $d{+}120$ \\
    Relation       & Date relation injected  & $d{+}30$, $d{+}120$ \\
    User-reference & Named reference date    & $d{-}15$, $d{+}30$ \\
    \bottomrule
  \end{tabular}
\end{table}

\subsection{Model routes and dataset coverage}

The evaluation dataset comprises 940 main responses collected across five model routes accessed through a unified API gateway: DeepSeek-V4-Flash, Qwen3.5-397B-A17B, hy3, DeepSeek-V4.1-Flash, and qwen3-vl-plus. The first four routes expose intermediate reasoning traces alongside their final answers, whereas qwen3-vl-plus returns only the final response. The 940 recorded responses are distributed across the five conditions, comprising 210 responses each in Natural and Delivered, 190 each in Forced and Relation, and 140 in User-reference. Completed runs cover 14 core service event types in Natural, Delivered, and User-reference, spanning scenarios such as pre-employment health checks, visa interviews, thesis defenses, and kindergarten admissions. Coverage expands to 19 event types in Forced and Relation by adding driving test, house viewing, immigration intake, language test, and medical exam scenarios. Alongside the 940 main responses, we collect 350 standalone responses for an explicit-date baseline and 187 runs for a multi-run stability audit. Across all analyses, denominators reflect responses with resolved labels for the specific observable under evaluation.

\subsection{Annotation protocol, scoring rules, and statistical testing}
\label{sec:annotation}

We combine rule-based extractors with condition-blind automated judges to score the behavioral observables defined in \S\ref{sec:observables}. The tool-call indicator $F$ is extracted directly from execution logs. The reasoning indicator $R$ uses a verified phrase matcher over exposed reasoning traces to detect explicit statements that the event date has passed. An auxiliary rule $A$ independently flags past-date statements in the final answer text. The response-level labels $S$ (upcoming framing), $T$ (past date position), and $P$ (advice acceptability) are assigned by a condition-blind GLM-5.3-flash judge. The judge receives the final answer, the scheduled event date, and the reference timestamp, and every assigned label requires a supporting verbatim quotation. Under the rubric, $P{=}1$ requires avoiding preparation for an elapsed appointment and providing appropriate evening guidance such as rest or clarifying which event the user meant.

To verify scoring reliability, an independent GPT-family judge blindly annotated 408 unambiguously matched responses across the Natural and Delivered conditions using the identical rubric. Cross-judge agreement reaches 366/379 resolved pairs for $S$ (Cohen's $\kappa = 0.928$), 392/393 for $T$ ($\kappa = 0.993$), and 370/391 for $P$ ($\kappa = 0.878$). In addition, comparing the judge label $S$ against the independent answer rule $A$ across 633 resolved post-date responses shows complementary labels in 605 cases (95.6\%). Because each route-by-profile unit contributes responses across multiple offsets and conditions, we evaluate paired transitions using exact two-sided McNemar tests over matched units. We apply Holm step-down adjustments within the three supplied-information comparisons and within the three hiatus-interval comparisons, complemented by per-route breakdowns and leave-one-route-out checks. All task prompts, episode profiles, scoring rubrics, and recorded outputs are archived in our supplementary benchmark package for offline reproduction.

\section{Results and Analysis}

\subsection{RQ1: recognizing temporal dependency and acquiring time}

We first verify baseline temporal competence through a standalone explicit-date test covering five fundamental operations across the 14 core event types, totaling 70 items per route and 350 responses overall. The five operations test 30-day interval arithmetic, past-date classification, future-date classification, and upcoming status judgments before and after the scheduled event. Four of the five routes (DeepSeek-V4-Flash, Qwen3.5-397B-A17B, hy3, and DeepSeek-V4.1-Flash) achieve near-perfect accuracy, resolving 69/70 or 70/70 items correctly (Appendix Table~\ref{tab:baseline}). The remaining route, qwen3-vl-plus, resolves 57/70 items overall, succeeding on all 14 interval, future-date, and explicit status items while resolving 1/14 past-date classification items. These baseline results confirm that the primary reasoning models possess strong calendar arithmetic capabilities when dates are presented explicitly.

In the main session-rekindling experiments, active clock acquisition responds strongly to whether the user prompt supplies the current time. Across the four routes with exposed reasoning, agents invoke the \Clock{} tool in 157/168 Natural responses (93.5\%) when the prompt omits the date. By contrast, they invoke the \Clock{} tool in only 30/168 Delivered responses (17.9\%) when the prompt already includes the timestamp. Across all five routes combined, \Clock{} calls occur in 158/210 Natural responses versus 30/210 Delivered responses. Focusing on post-date Natural episodes ($d{+}30$ and $d{+}120$ pooled, Table~\ref{tab:clock}), the four reasoning routes invoke the clock in 105 of 112 turns. Specifically, DeepSeek-V4-Flash calls \Clock{} in 27/28 turns, Qwen3.5-397B-A17B in 25/28, hy3 in 25/28, and DeepSeek-V4.1-Flash in 28/28. The qwen3-vl-plus route calls the clock once across 42 Natural turns and zero times post-date, mirroring its weaker baseline sensitivity on past-date items. When temporal cues are absent from a session-rekindling request, reasoning-enabled agents consistently recognize the temporal dependency and actively query the external clock.

\subsection{RQ3: advice appropriateness and outcome uncertainty}
\label{sec:rq3}

Under RQ3, we evaluate whether the agent's final advice fits the passed-unresolved state implied by the elapsed date. We examine the 88 post-date Natural responses where the agent called \Clock{} and obtained resolved labels for both $S$ and $P$. Within this set, the advice rubric accepts 41/47 non-upcoming responses (87.2\%) and rejects all 41 upcoming responses (0/41 accepted). Route-level acceptance in Table~\ref{tab:clock} mirrors upcoming-framing rates: DeepSeek-V4.1-Flash and hy3 achieve 18/20 and 17/22 accepted answers, whereas DeepSeek-V4-Flash and Qwen3.5-397B-A17B each achieve only 3/23. The six non-upcoming responses rejected by the rubric fail because they stop at asking which event the user meant rather than offering actionable evening guidance. Furthermore, across the 189 resolved post-date Relation responses where upcoming framing drops to zero ($S{=}0$), the lexical completion flag $K$ appears in 102/189 responses (54.0\%). Models frequently adopt completion-associated phrasing when deactivating an expired schedule instead of preserving open outcome uncertainty.

Qualitative inspection of the flagged post-date failures highlights three recurring behavioral patterns across model routes. The first pattern is retained-plan reuse, where the agent repeats the historical note to rest tonight and try again tomorrow almost verbatim despite past-date reasoning. In the June~5 health-check episode (Figure~\ref{fig:comic}), DeepSeek-V4-Flash notes that May~6 has passed, yet tells the user to copy the May~6 details and retry tomorrow. The second pattern is post-date preparation instruction, where the response provides a detailed checklist of documents and fasting rules for the elapsed appointment. The third pattern is clock-response detachment, where the agent executes a \Clock{} call at the start of the turn but generates advice that ignores the returned timestamp. In addition, our multi-run stability audit across 187 repeated generations over 48 selected evaluation units confirms that $S$ labels remain identical in 35/48 units. Counting all 140 planned post-date Natural turns bounds the upcoming share between 73/140 and 89/140 (52.1\% to 63.6\%), showing that stale framing is a systematic tendency.

\begin{table*}[t]
  \centering\small
  \caption{Post-date Natural responses by route showing clock invocations, upcoming framing, past-date wording in exposed reasoning, and rubric-accepted advice over resolved denominators.}
  \label{tab:clock}
  \begin{tabular}{@{}lccccc@{}}
    \toprule
    Route & Clock call & Upcoming with clock & Upcoming, clock omitted & Past-date wording & Advice accepted \\
    \midrule
    DeepSeek-V4-Flash   & 27  & 20/24 & 1/1  & 27/27  & 3/23 \\
    Qwen3.5-397B-A17B   & 25  & 19/23 & 3/3  & 24/25  & 3/23 \\
    hy3                 & 25  & 3/24  & 2/2  & 25/25  & 17/22 \\
    DeepSeek-V4.1-Flash & 28  & 0/21 & --   & 28/28  & 18/20 \\
    qwen3-vl-plus       & 0   & --    & 25/26 & --    & -- \\
    \midrule
    All routes          & 105 & 42/92 & 31/32 & 104/105 & 41/88 \\
    \bottomrule
  \end{tabular}
\end{table*}

\subsection{RQ2: updating task state and schedule framing}
\label{sec:rq2}

\subsubsection{Date-crossing comparisons}

\begin{figure*}[t]
\centering
\includegraphics[width=\textwidth]{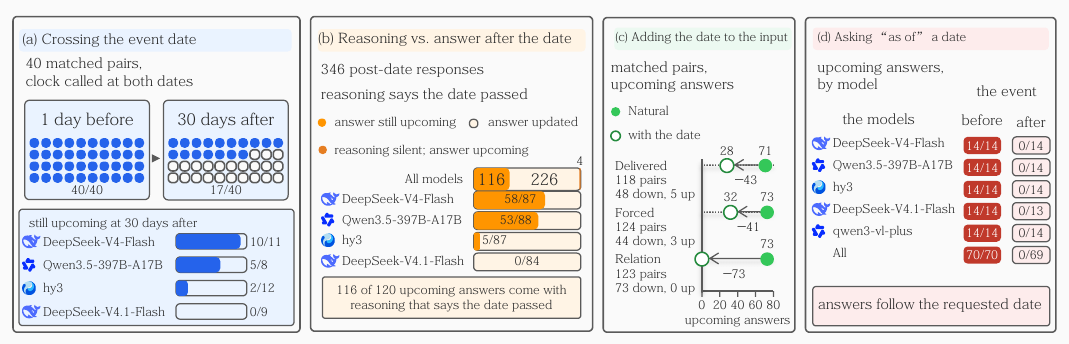}
\caption{Empirical results across conditions showing (a) pre-date versus post-date upcoming framing, (b) reasoning-answer dissociation, (c) reductions under supplied temporal inputs, and (d) adherence to user reference dates.}
\Description{A four-panel results figure: matched pre- and post-date upcoming framing, the reasoning-answer dissociation counts, upcoming-framing reductions under supplied temporal inputs, and user-reference adherence at two offsets.}
\label{fig:conditions}
\end{figure*}

To evaluate whether agents update their task state once a scheduled date passes, we compare matched Natural responses at $d{-}1$ versus $d{+}30$. This comparison covers 40 route-by-profile pairs where the agent called \Clock{} and obtained resolved $S$ labels at both time points. One day before the appointment ($d{-}1$), all 40 responses appropriately frame the event as upcoming (Figure~\ref{fig:conditions}a). Thirty days later ($d{+}30$), 23/40 responses drop upcoming framing (exact McNemar $p=2.38\times 10^{-7}$), while 17/40 still frame the expired schedule as upcoming. Across all resolved post-date Natural responses with a \Clock{} call, 42/92 responses retain upcoming framing, alongside 31/32 responses where \Clock{} was omitted (Table~\ref{tab:clock}).

Susceptibility to post-date upcoming framing varies substantially across model routes. Within the 40 matched $d{-}1$ versus $d{+}30$ pairs, DeepSeek-V4-Flash retains upcoming framing in 10/11 post-date pairs and Qwen3.5-397B-A17B retains it in 5/8 pairs. By contrast, hy3 retains upcoming framing in 2/12 pairs and DeepSeek-V4.1-Flash drops it in all 9/9 pairs. Extending the conversational hiatus to four months ($d{+}120$) yields a similar error rate, with 18/39 matched post-date responses still framing the passed schedule as upcoming. Together, the $d{+}30$ and $d{+}120$ comparisons span 47 distinct route-by-profile units and show that active clock retrieval frequently fails to deactivate obsolete preparation plans.

\subsubsection{Dissociation between reasoning and final advice}

We next examine whether post-date upcoming framing stems from failing to notice that the scheduled date has elapsed during intermediate reasoning. Pooling post-date responses with exposed reasoning and resolved $S$ labels across Natural (with \Clock{} calls), Delivered, and Forced yields 346 responses across the four reasoning routes. This sample comprises 87 responses for DeepSeek-V4-Flash, 88 for Qwen3.5-397B-A17B, 87 for hy3, and 84 for DeepSeek-V4.1-Flash. Across these 346 post-date traces, 342 contain explicit past-date wording ($R{=}1$) acknowledging that the scheduled date has already passed (Figure~\ref{fig:conditions}b). Remarkably, 120 of the 346 responses still adopt upcoming framing ($S{=}1$) in their final answer, and 116 of those 120 failures satisfy the co-occurrence diagnostic $U{=}1$ (Eq.~\ref{eq:cooc}). Even under the core phrase lexicon prior to the three documented phrase additions, 113 of the 120 upcoming responses already carry past-date reasoning wording. DeepSeek-V4-Flash accounts for 58/87 of the $U{=}1$ occurrences and Qwen3.5-397B-A17B accounts for 53/88, with hy3 contributing 5/87 and DeepSeek-V4.1-Flash contributing 0/84. In all 116 episodes, the model explicitly deduces in its chain of thought that the appointment date has passed, yet still instructs the user to prepare for that appointment.

\subsubsection{Impact of supplied temporal information and reference dates}

Matched post-date comparisons demonstrate that strengthening the prominence and structure of temporal inputs significantly reduces stale upcoming framing (Figure~\ref{fig:conditions}c). Supplying the current date directly in the prompt (Natural $\to$ Delivered) reduces upcoming framing from 71 to 28 across 118 matched pairs. This shift represents a 36.4 percentage-point reduction (48 downward versus 5 upward transitions, Holm-adjusted McNemar $p=7.08\times 10^{-10}$). Injecting the clock observation before generation (Natural $\to$ Forced) lowers upcoming framing from 73 to 32 across 124 pairs, a 33.1-point drop (Holm-adjusted $p=4.93\times 10^{-10}$). Providing the computed date relation (Natural $\to$ Relation) eliminates upcoming framing completely, dropping from 73 to 0 across 123 matched pairs (Holm-adjusted $p=6.35\times 10^{-22}$).

This reduction holds consistently across deployments, as every route with baseline errors improves under Delivered, Forced, and Relation (Appendix Table~\ref{tab:routepairs}). Furthermore, leave-one-route-out checks preserve the direction and significance of all three contrasts. Under Relation, upcoming framing drops to 0/37 for DeepSeek-V4-Flash and 0/38 for each of the other four routes. Similarly, under the User-reference condition (Figure~\ref{fig:conditions}d), all five routes follow the requested perspective with 70/70 upcoming responses at $d{-}15$ and 0/69 at $d{+}30$. When the temporal anchor or date relation is foregrounded as an explicit task constraint, models reliably deactivate the outdated preparation plan.

\subsubsection{Hiatus interval sensitivity}

Comparing matched post-date responses between a 30-day hiatus ($d{+}30$) and a 120-day hiatus ($d{+}120$) reveals that a longer elapsed interval fails to mitigate stale framing. Under Forced, upcoming framing rises from 18/93 at $d{+}30$ to 30/93 at $d{+}120$ (13 upward versus 1 downward transition, Holm-adjusted McNemar $p\approx 0.005$). Under the Delivered condition, upcoming framing shifts from 11/62 at $d{+}30$ to 19/62 at $d{+}120$ (14 upward versus 6 downward transitions, unadjusted $p=0.1153$). Meanwhile, the Natural condition remains stable at 34/55 at $d{+}30$ versus 35/55 at $d{+}120$ ($p=1.0$). Rather than making the expiration of an earlier appointment more obvious, a four-month conversational hiatus can make models even more prone to reviving historical preparation plans.

\begin{table*}[t]
  \centering\small
  \caption{Session-rekindling benchmark leaderboard across the five evaluated model routes showing percentage scores and resolved counts across the six diagnostic metrics.}
  \label{tab:leaderboard}
  \setlength{\tabcolsep}{5pt}
  \begin{tabular}{@{}clcccccc@{}}
    \toprule
    Rank & Model Route & Base$\uparrow$ & Clock acq. ($F$) $\uparrow$ & Past reason ($R$) $\uparrow$ & Stale frame ($S$) $\downarrow$ & Dissociation ($U$) $\downarrow$ & Advice acc. ($P$) $\uparrow$ \\
    \midrule
    1 & DeepSeek-V4.1-Flash & 98.6\% (69/70)  & 100.0\% (28/28) & 100.0\% (84/84) & \textbf{0.0\%} (0/21)  & \textbf{0.0\%} (0/84)   & \textbf{90.0\%} (18/20) \\
    2 & hy3                 & 100.0\% (70/70) & 89.3\% (25/28)  & 96.6\% (84/87)  & 19.2\% (5/26)  & 5.7\% (5/87)    & 77.3\% (17/22) \\
    3 & DeepSeek-V4-Flash   & 100.0\% (70/70) & 96.4\% (27/28)  & 98.9\% (86/87)  & 84.0\% (21/25) & 66.7\% (58/87)  & 13.0\% (3/23) \\
    4 & Qwen3.5-397B-A17B   & 100.0\% (70/70) & 89.3\% (25/28)  & 100.0\% (88/88) & 84.6\% (22/26) & 60.2\% (53/88)  & 13.0\% (3/23) \\
    5 & qwen3-vl-plus       & 81.4\% (57/70)  & 0.0\% (0/28)    & --              & 96.2\% (25/26) & --              & -- \\
    \midrule
    --& Overall             & 96.0\% (336/350)& 75.0\% (105/140)& 98.8\% (342/346)& 58.9\% (73/124)& 33.5\% (116/346) & 46.6\% (41/88) \\
    \bottomrule
  \end{tabular}
\end{table*}

\section{The Session-Rekindling Benchmark}
\label{sec:benchmark}

Building on the empirical findings from RQ1 through RQ3, we package our controlled task episodes, virtual-clock harness, and diagnostic scoring pipeline into a reusable session-rekindling benchmark. The benchmark comprises 940 main evaluation cells across the five temporal-information conditions, 350 standalone explicit-date baseline items covering five calendar operations, and 48 multi-run stability units. For any evaluated agent, the scoring script computes six lightweight rates directly from the response logs. It measures explicit-date accuracy, post-date clock acquisition ($F{=}1$), past-date reasoning ($R{=}1$), and advice acceptance ($P{=}1$), alongside post-date stale framing ($S{=}1$) and dissociation ($U{=}1$). Table~\ref{tab:leaderboard} presents the benchmark leaderboard across all five evaluated model routes, ranked by post-date advice acceptance and stale framing suppression.

The leaderboard highlights three diagnostic regimes across current model routes. First, DeepSeek-V4.1-Flash and hy3 achieve strong adaptation, pairing high clock acquisition with low stale framing ($0.0\%$ and $19.2\%$) and high advice acceptance ($90.0\%$ and $77.3\%$). Second, DeepSeek-V4-Flash and Qwen3.5-397B-A17B exemplify the knowing-doing dissociation. Both routes achieve near-perfect explicit-date accuracy, clock acquisition, and past-date reasoning ($98.9\%$ to $100.0\%$), yet suffer over $84\%$ stale framing and only $13.0\%$ advice acceptance. Third, qwen3-vl-plus fails at the initial time-acquisition stage ($0.0\%$ post-date clock calls), resulting in $96.2\%$ stale framing. The benchmark archive includes all episode profiles, prompt templates, virtual-clock specifications, condition-blind judge rubrics, and verification scripts for offline reproduction.

\section{Discussion}
\label{sec:explanations}

Our results reveal a persistent gap between knowing the current date and updating a retained plan. Three mechanisms account for why agents revive expired schedules during session rekindling. First, deictic re-anchoring can bind relative temporal terms such as \emph{tonight} and \emph{tomorrow} in retained notes directly to the follow-up turn, matching the rise in upcoming framing from 18/93 at 30 days to 30/93 at 120 days under Forced. Second, helpfulness-driven action inertia pressures instruction-tuned models to satisfy open-ended evening planning prompts by falling back on the concrete preparation checklist stored in memory. Third, procedural detachment between tool-calling heads and response planning allows models to emit clock calls as a routine prefix while relying on historical context momentum during generation.

Several scope boundaries inform the interpretation of our study and benchmark release. First, our suite evaluates synthetic Chinese service episodes via commercial API routes, so exposed reasoning indicators ($R$ and $U$) apply specifically to routes that return chain-of-thought tokens. Second, while our dual-judge protocol achieves high agreement ($\kappa \ge 0.878$) and 95.6\% complementarity with lexical rules, automated scoring complements rather than replaces human evaluation. Third, the lexical flag $K$ captures surface completion vocabulary including hedged mentions, motivating semantic entailment checks in future work.

\section{Conclusion}

This study investigated whether long-horizon agents adapt their recommendations when users rekindle a service conversation after a scheduled event date has passed. Experiments across 940 responses and five model routes show that agents reliably query the virtual clock when prompts omit the date, yet often still frame elapsed appointments as upcoming. Crucially, exposed reasoning states explicitly in 116 of 346 post-date cases that the scheduled date has passed while the final answer still recommends the outdated plan. Supplying explicit temporal context substantially reduces these errors, and externalizing the date-relation comparison eliminates upcoming framing completely while exposing a secondary tendency toward unverified completion phrasing. Our released session-rekindling benchmark and evaluation protocol provide a foundation for building temporally aware long-horizon agents across web platforms.

\bibliographystyle{ACM-Reference-Format}
\bibliography{refs}

@inproceedings{maharana2024longterm,
  author = {Maharana, Adyasha and Lee, Dong-Ho and Tulyakov, Sergey and Bansal, Mohit and Barbieri, Francesco and Fang, Yuwei},
  title = {Evaluating Very Long-Term Conversational Memory of {LLM} Agents},
  booktitle = {Proceedings of the 62nd Annual Meeting of the Association for Computational Linguistics (Volume 1: Long Papers)},
  pages = {13851--13870},
  year = {2024},
  doi = {10.18653/v1/2024.acl-long.747}
}

@inproceedings{wu2025longmemeval,
  author = {Wu, Di and Wang, Hongwei and Yu, Wenhao and Zhang, Yuwei and Chang, Kai-Wei and Yu, Dong},
  title = {{LongMemEval}: Benchmarking Chat Assistants on Long-Term Interactive Memory},
  booktitle = {International Conference on Learning Representations (ICLR)},
  year = {2025},
  note = {arXiv:2410.10813}
}

@misc{cheng2025temporally,
  author = {Cheng, Yize and others},
  title = {Your {LLM} Agents are Temporally Blind: The Misalignment Between Tool Use Decisions and Human Time Perception},
  year = {2025},
  note = {arXiv:2510.23853}
}

@misc{chao2026stale,
  author = {Chao, Hanxiang and Bai, Yihan and Sheng, Rui and Li, Tianle and Sun, Yushi},
  title = {{STALE}: Can {LLM} Agents Know When Their Memories Are No Longer Valid?},
  year = {2026},
  note = {arXiv:2605.06527}
}

@misc{liu2026pmbench,
  author = {Liu, Genglin and Gabriel, Saadia},
  title = {{PM-Bench}: Evaluating Prospective Memory in {LLM} Agents},
  year = {2026},
  note = {arXiv:2607.12385}
}

@misc{zhang2026trigger,
  author = {Zhang, Tianhua and Wang, Xinjiang and Zhang, Qianxi and Chen, Qi and Li, Kun and Chen, Yaoqi and Wang, Dingdong and Meng, Helen and Lu, Yan},
  title = {{TriggerBench}: Investigating Prospective Memory for Large Language Models},
  year = {2026},
  note = {arXiv:2606.23459}
}

@misc{wang2026contract,
  author = {Wang, Jicheng and He, Yifeng and Wang, Zili and Xing, Hanwen and De, Arkaprava and Chen, Hao},
  title = {{ContractBench}: Can {LLM} Agents Preserve Observation Contracts?},
  year = {2026},
  note = {arXiv:2605.17281}
}

@misc{fan2026track,
  author = {Fan, Xinyi and Liu, Miri and Yang, Ruozhen and Ouyang, Siru and Han, Jiawei},
  title = {Can Agent Memory Systems Track Evolving State?},
  year = {2026},
  note = {arXiv:2608.19652}
}

@misc{sun2026repair,
  author = {Sun, Haofei and He, Lin},
  title = {When Memory Updates but Behavior Does Not: Repairing Implicit Stale Dependencies in Personalized Agent Responses},
  year = {2026},
  note = {arXiv:2608.01619}
}

@misc{lanham2023faithfulness,
  author = {Lanham, Tamera and Chen, Anna and Radhakrishnan, Ansh and others},
  title = {Measuring Faithfulness in Chain-of-Thought Reasoning},
  year = {2023},
  note = {arXiv:2307.13702}
}

@article{moschl2020aftereffects,
  author = {M{\"o}schl, Marcus and Fischer, Rico and Bugg, Julie M. and Scullin, Michael K. and Goschke, Thomas and Walser, Moritz},
  title = {Aftereffects and Deactivation of Completed Prospective Memory Intentions: A Systematic Review},
  journal = {Psychological Bulletin},
  volume = {146},
  number = {3},
  pages = {245--278},
  year = {2020},
  doi = {10.1037/bul0000221}
}

@inproceedings{das2026time,
  title = {TIME: Temporally Intelligent Meta-reasoning Engine for Context-Triggered Explicit Reasoning},
  author = {Association for Computational Linguistics 2026 and Susmit Das},
  year = {2026},
  journal = {Underline Science Inc.},
  booktitle = {Underline Science Inc.}
}

@misc{herel2025timeaware,
  author = {Herel, David and Bartek, Vojtech and Jirak, Jiri and Mikolov, Tomas},
  title = {Time Awareness in Large Language Models: Benchmarking Fact Recall Across Time},
  year = {2025},
  note = {arXiv:2409.13338}
}

@inproceedings{deng2023mind2web,
  author = {Deng, Xiang and Gu, Yu and Zheng, Boyuan and Chen, Shijie and Stevens, Samuel and Wang, Boshi and Sun, Huan and Su, Yu},
  title = {{Mind2Web}: Towards a Generalist Agent for the Web},
  booktitle = {Advances in Neural Information Processing Systems 36 (Datasets and Benchmarks Track)},
  year = {2023}
}

@inproceedings{lu2024weblinx,
  title = {WebLINX: Real-World Website Navigation with Multi-Turn Dialogue},
  author = {Siva Reddy and Xing Lu and Zden{\v{e}}k Kasner},
  year = {2024},
  journal = {Institute of Formal and Applied Linguistics (ÚFAL)},
  booktitle = {Institute of Formal and Applied Linguistics (ÚFAL)}
}

@inproceedings{zhou2024webarena,
  author = {Zhou, Shuyan and Xu, Frank F. and Zhu, Hao and Zhou, Xuhui and Lo, Robert and Sridhar, Abishek and Cheng, Xianyi and Bisk, Yonatan and Fried, Daniel and Alon, Uri and Neubig, Graham},
  title = {{WebArena}: A Realistic Web Environment for Building Autonomous Agents},
  booktitle = {International Conference on Learning Representations (ICLR)},
  year = {2024}
}

@inproceedings{drouin2024workarena,
  author = {Drouin, Alexandre and Gasse, Maxime and Caccia, Massimo and Laradji, Issam H. and Del Verme, Manuel and Marty, Tom and Boisvert, L{\'e}o and Thakkar, Megh and Cappart, Quentin and Vazquez, David and Chapados, Nicolas and Lacoste, Alexandre},
  title = {{WorkArena}: How Capable are {Web} Agents at Solving Common Knowledge Work Tasks?},
  booktitle = {Proceedings of the 41st International Conference on Machine Learning},
  series = {PMLR},
  volume = {235},
  pages = {11642--11662},
  year = {2024}
}

@inproceedings{xu2025agentcompany,
  author = {Xu, Frank F. and Song, Yufan and Li, Boxuan and Tang, Yuxuan and Jain, Kritanjali and Bao, Mengxue and Wang, Zora Zhiruo and Zhou, Xuhui and Guo, Zhitong and Cao, Murong and Yang, Mingyang and Lu, Hao Yang and Martin, Amaad and Su, Zhe and Maben, Leander and Mehta, Raj and Chi, Wayne and Jang, Lawrence Keunho and Xie, Yiqing and Zhou, Shuyan and Neubig, Graham},
  title = {{TheAgentCompany}: Benchmarking {LLM} Agents on Consequential Real World Tasks},
  booktitle = {Advances in Neural Information Processing Systems 38 (Datasets and Benchmarks Track)},
  year = {2025}
}

@article{lesellier2025browsergym,
  author = {Le Sellier de Chezelles, Thibault and Gasse, Maxime and Lacoste, Alexandre and Drouin, Alexandre and Caccia, Massimo and Boisvert, L{\'e}o and Thakkar, Megh and Marty, Tom and Assouel, Rim and Shayegan, Sahar Omidi and Jang, Lawrence Keunho and L{\`u}, Xing Han and Yoran, Ori and Kong, Dehan and Xu, Frank F. and Reddy, Siva and Cappart, Quentin and Neubig, Graham and Salakhutdinov, Ruslan and Chapados, Nicolas},
  title = {The {BrowserGym} Ecosystem for Web Agent Research},
  journal = {Transactions on Machine Learning Research},
  year = {2025},
  issn = {2835-8856}
}

@inproceedings{yoran2024assistantbench,
  author = {Yoran, Ori and Amouyal, Samuel Joseph and Malaviya, Chaitanya and Bogin, Ben and Press, Ofir and Berant, Jonathan},
  title = {{AssistantBench}: Can Web Agents Solve Realistic and Time-Consuming Tasks?},
  booktitle = {Proceedings of EMNLP 2024},
  pages = {8938--8968},
  year = {2024},
  doi = {10.18653/v1/2024.emnlp-main.505}
}

\appendix

\section{Task Inputs and Experimental Protocol}
\label{app:inputs}

Each episode uses the standardized Chinese probe ``\begin{CJK}{UTF8}{gbsn}结合你之前了解的情况，你觉得我今晚应该怎么安排？\end{CJK}''. This prompt translates to asking how the user should plan tonight based on prior context. The retained session context provides four structured records: the target booking confirmation, preparation notes, an unrelated secondary invitation, and an earlier plan to rest tonight and resume tomorrow. The Natural condition presents the base probe with the \Clock{} tool available. The Delivered condition prepends the current date and time to the probe, while the Forced condition injects a completed \Clock{} tool call and its returned timestamp before the assistant turn. The Relation condition injects a structured time-context note stating the current date, the scheduled event date, and whether the scheduled date has already passed. The User-reference condition asks the agent to answer the base probe as of an explicitly stated date ($d{-}15$ or $d{+}30$) with the virtual clock matching that date.

All benchmark artifacts preserve the exact prompt templates, episode profiles, and decoding configurations. The explicit-date baseline evaluates all routes at temperature 0.0 to measure deterministic date comparison accuracy. The main evaluations record responses under each route's standard configuration, and our 187-run audit confirms that framing remains stable in 35/48 units. All reported statistics and tables are reproduced offline from the archived response logs through deterministic verification scripts in the supplementary package.

\section{Scoring Rubric and Reliability Audit}
\label{app:rules}

The automated evaluation protocol scores each response using explicit rubric definitions and verbatim evidence quotes. The upcoming-framing label $S$ equals 1 when the response offers preparation steps, reminders, or reassurance that treats the scheduled event as still ahead, and 0 when it drops upcoming treatment. The date-position label $T$ records whether the final answer places the scheduled event in the past ($T{=}1$) or in the future ($T{=}0$). The advice-acceptability label $P$ equals 1 when the response avoids preparing for an elapsed appointment and provides appropriate evening guidance, such as resting or clarifying the status of past events. The lexical completion flag $K$ detects vocabulary associated with event completion in the final answer, while the answer rule $A$ detects explicit past-date statements in the final response text.

The reasoning matcher $R$ scans exposed chain-of-thought traces for explicit phrases stating that the scheduled date has already elapsed. Beyond the core phrase lexicon, the matcher incorporates three documented phrase additions covering expressions observed in two DeepSeek-V4-Flash Delivered responses and one Qwen3.5-397B-A17B Forced response. In the dual-judge reliability audit, the primary GLM-5.3-flash judge and the secondary GPT-family judge independently evaluated 408 unambiguously matched responses spanning the Natural and Delivered conditions. The resulting $2\times 2$ contingency tables confirm strong inter-judge reliability across all three primary metrics. Specifically, the two judges achieve 366/379 agreement on $S$ ($\kappa = 0.928$), 392/393 on $T$ ($\kappa = 0.993$), and 370/391 on $P$ ($\kappa = 0.878$).

\section{Additional Empirical Results}
\label{app:results}

Table~\ref{tab:baseline} reports route-level accuracy on the 350-response explicit-date baseline across the five date-reasoning operations. DeepSeek-V4-Flash, Qwen3.5-397B-A17B, and hy3 resolve all 70 items across the 14 core event types, and DeepSeek-V4.1-Flash resolves 69/70 items. The qwen3-vl-plus route resolves all 14 interval arithmetic items, future-date items, and status items when prompted directly with both dates, while resolving 1/14 past-date classification items. Across the main session-rekindling experiments, pooling post-date responses at $d{+}30$ and $d{+}120$ yields resolved upcoming-framing counts ($S{=}1$) of 73/124 in Natural and 32/132 in Delivered. In the remaining post-date conditions, resolved upcoming-framing counts reach 48/188 in Forced and 0/189 in Relation (Table~\ref{tab:routecond}). Meanwhile, one day prior to the event ($d{-}1$), all 65/65 resolved Natural responses and 63/63 resolved Delivered responses appropriately adopt upcoming framing.

\begin{table}[ht]
  \centering\small\setlength{\tabcolsep}{4pt}
  \caption{Explicit-date baseline accuracy on the five temporal operations over the 14 core event types: 30-day interval arithmetic (Interval), past- and future-date classification (Past, Future), and upcoming-status judgments before and after the scheduled event (Before, After).}
  \label{tab:baseline}
  \begin{tabular}{@{}lccccc@{}}
    \toprule
    Route & Interval & Past & Future & Before & After \\
    \midrule
    DeepSeek-V4-Flash   & 14/14 & 14/14 & 14/14 & 14/14 & 14/14 \\
    Qwen3.5-397B-A17B   & 14/14 & 14/14 & 14/14 & 14/14 & 14/14 \\
    hy3                 & 14/14 & 14/14 & 14/14 & 14/14 & 14/14 \\
    DeepSeek-V4.1-Flash & 14/14 & 14/14 & 14/14 & 14/14 & 13/14 \\
    qwen3-vl-plus       & 14/14 &  1/14 & 14/14 & 14/14 & 14/14 \\
    \bottomrule
  \end{tabular}
\end{table}

For the date-position metric $T$, post-date responses place the scheduled date in the past ($T{=}1$) in 54/132 Natural responses and 110/134 Delivered responses. Under Forced and Relation, post-date $T{=}1$ counts reach 143/190 and 187/187 respectively. Comparing the upcoming-framing judge label $S$ with the independent answer rule $A$ across the 633 resolved post-date responses yields 605 complementary pairs. These 605 pairs consist of 468 responses with $S{=}0, A{=}1$ and 137 responses with $S{=}1, A{=}0$. The remaining 28 responses include 12 cases with $S{=}0, A{=}0$ and 16 cases with $S{=}1, A{=}1$. In those 16 cases, the final answer states that the date has passed yet still recommends preparation. Table~\ref{tab:routepairs} details the route-level matched post-date comparisons between Natural and each supplied-information condition, confirming that every route with baseline errors improves under Delivered, Forced, and Relation.

\begin{table}[ht]
  \centering\small
  \caption{Post-date upcoming framing counts ($S{=}1$) by model route and condition, pooling $d{+}30$ and $d{+}120$ over resolved responses.}
  \label{tab:routecond}
  \begin{tabular}{@{}lcccc@{}}
    \toprule
    Route & Natural & Delivered & Forced & Relation \\
    \midrule
    DeepSeek-V4-Flash   & 21/25 & 11/25 & 27/38 & 0/37 \\
    Qwen3.5-397B-A17B   & 22/26 & 17/27 & 21/38 & 0/38 \\
    hy3                 &  5/26 &  2/26 &  0/37 & 0/38 \\
    DeepSeek-V4.1-Flash &  0/21 &  0/26 &  0/37 & 0/38 \\
    qwen3-vl-plus       & 25/26 &  2/28 &  0/38 & 0/38 \\
    \midrule
    All routes          & 73/124 & 32/132 & 48/188 & 0/189 \\
    \bottomrule
  \end{tabular}
\end{table}

\begin{table}[ht]
  \centering\small\setlength{\tabcolsep}{4pt}
  \caption{Matched post-date upcoming framing transitions from Natural to each supplied-information condition by route, with total matched pairs in parentheses.}
  \label{tab:routepairs}
  \begin{tabular}{@{}lccc@{}}
    \toprule
    Route & Delivered & Forced & Relation \\
    \midrule
    DeepSeek-V4-Flash   & 20$\to$11 (23) & 21$\to$19 (25) & 21$\to$0 (24) \\
    Qwen3.5-397B-A17B   & 21$\to$15 (25) & 22$\to$13 (26) & 22$\to$0 (26) \\
    hy3                 &  5$\to$1 (24)  &  5$\to$0 (26)  &  5$\to$0 (26) \\
    DeepSeek-V4.1-Flash &  0$\to$0 (20)  &  0$\to$0 (21)  &  0$\to$0 (21) \\
    qwen3-vl-plus       & 25$\to$1 (26)  & 25$\to$0 (26)  & 25$\to$0 (26) \\
    \bottomrule
  \end{tabular}
\end{table}

\section{Qualitative Session-Rekindling Case Study}
\label{app:case}

We illustrate the reasoning-answer dissociation with a representative post-date Natural episode from the pre-employment medical check-up profile evaluated on DeepSeek-V4-Flash (depicted in Figure~\ref{fig:comic}). In this episode, the initial conversation took place on March~1, 2026, regarding a health check scheduled for May~6, 2026, at 9~AM. The retained notes also record the user's earlier plan to rest tonight and try again tomorrow. When the user rekindles the session on June~5, 2026, and asks how to plan for the evening, the agent invokes \Clock{} and receives the June~5 timestamp. In its exposed reasoning trace, the model explicitly notes the May~6 appointment date and states that this date is already one month in the past.

Despite recognizing that the appointment date passed a month earlier, the model's final answer revives the March preparation plan almost verbatim. In translation, the response advises the user to write down the key details on paper, including the May~6 at 9~AM appointment time and the confirmation form. It further instructs the user to verify their name, ID number, and clinic location before sleeping early to try again tomorrow. The condition-blind judge assigns $S{=}1$, $T{=}0$, and $P{=}0$, yielding $U{=}1$ together with $R{=}1$. Evaluated on the identical episode and June~5 clock reading, DeepSeek-V4.1-Flash notes in its reasoning that the May~6 check-up has long passed. It then drops the obsolete preparation plan in its final response ($S{=}0$, $T{=}1$, $P{=}1$) and asks which upcoming event the user would like to plan for tonight.

\end{document}